\documentclass[runningheads]{llncs}
\usepackage[T1]{fontenc}
\usepackage{graphicx}
\usepackage{comment}
\usepackage{varwidth}
\usepackage{multirow}
\usepackage{array}
\usepackage{float}
\usepackage{amsmath}
\usepackage{graphicx}
\usepackage{subcaption}
\usepackage[section]{placeins}

\begin{document}
%
%\title{Automated Evaluation of ChatInvent: Using LLM-as-a-Judge for Evaluation of Agentic Drug Discovery}
\title{Designing a Robust LLM-Based Evaluation System for Agentic AI in Drug Discovery Through Human Alignment}
\titlerunning{Designing LLM-Based Evaluation System for Agentic AI in Drug Discovery}
\author{Emma Granqvist\inst{1,2}\orcidID{0009-0000-6436-4178} \and
Rocío Mercado\inst{2,3}\orcidID{0000-0002-6170-6088} \and
Samuel Genheden\inst{1}\orcidID{0000-0002-7624-7363}}

\authorrunning{E. Granqvist et al.}
\institute{Molecular AI, Discovery Sciences, R\&D, AstraZeneca, Gothenburg, Sweden %\email{\{emma.granqvist,samuel.genheden\}@astrazeneca.com}
\and
Department of Computer Science and Engineering, Section for Data Science and AI, Chalmers University of Technology and University of Gothenburg, Gothenburg, Sweden \\
%\email{rocio.mercado@chalmers.se} 
\and
Science for Life Laboratory (SciLifeLab), Gothenburg, Sweden
}
\maketitle              % typeset the header of the contribution
\begin{abstract}
Agentic large language model (LLM) systems are reshaping scientific workflows in chemistry and drug discovery, but evaluating their open-ended, tool-augmented outputs remains a fundamental bottleneck. The LLM-as-a-Judge paradigm has emerged as a scalable alternative, but existing drug discovery benchmarks deploy LLM judges without validating their alignment with human experts. In this work, we present an LLM-as-a-Judge evaluation framework for ChatInvent, an agentic drug discovery assistant deployed at AstraZeneca, with five contributions. First, we define four output-quality evaluation dimensions---Completeness, Relevancy, Structural Clarity, and Scope Adherence---alongside deterministic Tool Call Correctness checks. Second, we validate the judge through a human alignment study with five expert annotators, comparing Gemini 3.1 Pro, Claude Opus 4.7, GPT-5, and Llama 3.1 70B as candidate judges. Third, we optimize the best-performing judge using few-shot demonstrations of human-annotated examples, improving alignment with the human majority vote from 0.80 to 0.86. Fourth, applying the optimized judge to 70 held-out questions, we surface concrete limitations and find no strong evidence that informal phrasing degrades output quality; it may, however, still be helpful to have the LLM rewrite the original question before querying the agent. Finally, we extend the framework to 38 adversarial questions that are ambiguous, invalid, out-of-scope or ethically sensitive, and show that the agent's refusal behavior is guided by the stated intent of a request. Our framework provides a reusable template for human-aligned evaluation of agentic systems in scientific domains.

\end{abstract}

\section{Introduction}

The emergence of large language models (LLMs) has transformed a wide range of domains due to their outstanding generalization, adaptability, and transfer capabilities \cite{naveed2025comprehensive}. In chemistry and drug discovery applications, LLM-based agentic systems have become powerful orchestrators by integrating external domain-specific tools and databases \cite{gottweis2025coscientist,m2024chemcrow,he2026democratising,mcnaughton2024cactus,ding2025scitoolagent,zhang2026molclaw}. This tool-augmented design allows LLM agents to leverage decades of computational chemistry software and scientific infrastructure while mitigating LLM limitations on tasks that require precise calculations, symbolic reasoning, or access to up-to-date knowledge. Agentic systems such as the AI co-scientist \cite{gottweis2025coscientist}, ChemCrow\cite{m2024chemcrow}, CACTUS \cite{mcnaughton2024cactus}, SciToolAgent\cite{ding2025scitoolagent}, MolClaw\cite{zhang2026molclaw}, and ChatInvent\cite{he2026democratising} illustrate this paradigm: they have the capability to plan experiments, call specialized computational tools, and coordinate multi-step scientific workflows, demonstrating how LLMs coupled with expert tools can support end-to-end scientific planning and execution.

The development of LLM powered frameworks also presented a pressing challenge for evaluation. Due to their powerful generative capabilities, these systems often produce open-ended and context-dependent output, making traditional metrics that compare predictions to a single ground truth insufficient for a comprehensive evaluation of reliability, usefulness, and robustness. Instead, human annotations can be regarded as the ``ground truth'' as these can offer a deeper understanding of the models performance. However, gathering human feedback from experts is typically time-consuming and resource intensive, making large-scale evaluation a challenge. As a result, LLMs have increasingly been used as evaluators by adapting their evaluation based on the task's context and criteria. The LLM-as-a-Judge approach enables scalable and autonomous evaluation while raising concerns about reliability, bias, and reproducibility, underscoring the need to also assess the quality of the evaluator itself.

In this work, we showcase how LLM-as-a-Judge can be designed and used for a reliable, automated, and scalable evaluation of an agentic drug discovery system, ChatInvent. Our contributions include:
\begin{itemize}
    \item an analysis of the design of an LLM-as-a-Judge for drug discovery systems,
    \item a study on human alignment of a multi-component LLM judge,
    \item a comparison of performance across different models underlying the LLM judge,
    \item an assessment of the quality of the agentic output across the different functionalities of ChatInvent,
    \item and an assessment of how the agent handles adversarial questions.
\end{itemize}

We focus on a specific agent, ChatInvent, rather than trying to compare different agents and we focus on introducing concepts that we believe are important when designing LLM-as-a-Judge systems for chemistry agents because the capability and scope of any agentic system constantly evolves.

\section{Related Work}

Traditional reference-based metrics such as BLEU \cite{papineni2002bleu} and ROUGE \cite{lin2004rouge} fail to capture the semantic nuances, task diversity, and open-ended nature of modern generative model outputs. Human preference is widely considered the gold-standard for evaluating LLM systems, but expert evaluation does not scale to the iteration speed of modern agent development, creating a bottleneck for rapid prototyping and large-batch assessment.

The LLM-as-a-Judge paradigm, introduced by Zheng et al.~\cite{zheng2023judging}, addresses this by using a strong LLM to evaluate model outputs at scale while maintaining high agreement with human judgment. Strong LLM judges can reach over 80\% agreement with human evaluators on general tasks~\cite{zheng2023judging,jung2025trust}, matching or exceeding the agreement between individual human annotators at a fraction of the time. However, LLM judges exhibit several systematic biases, including model self-preference, format and verbosity bias, and positional bias~\cite{zheng2023judging},  underscoring the importance of validating the judge itself against human annotations rather than treating its outputs as ground truth.

Recent work has framed LLM-based evaluation and prompt optimization as complementary tools to build scalable evaluation systems. DSPy~\cite{khattab2024dspy} provides a programming model in which evaluation criteria are expressed as typed signatures that can be optimized via demonstrations, which we adopt as the implementation substrate for our judge. Most directly related to our setting, Ríos-García and Jablonka~\cite{rios-garcia2025llmjudge} use an LLM-as-judge and LLM-as-optimizer framework for organic chemistry data extraction, demonstrating that lightweight prompt refinement against expert annotations can substantially improve domain-specific evaluation quality.

For agentic drug discovery systems specifically, two comprehensive evaluation efforts have been described. SciToolEval was developed to benchmark SciToolAgent~\cite{ding2025scitoolagent}, a single-agent orchestrator, using LLM-generated test questions covering both single- and multi-tool workflows; an LLM judge scores both the final answer and the tool-call sequence, and the benchmark has been used to compare SciToolAgent against ChemCrow~\cite{m2024chemcrow} and CACTUS~\cite{mcnaughton2024cactus}. MolClaw~\cite{zhang2026molclaw} introduces MolBench (based on ChemCoTBench), a multi-dimensional benchmark spanning diverse chemistry tasks and evaluating performance against both stand-alone LLMs and other agentic frameworks along dimensions such as tool-call sequence and scientific validity. In contrast to these efforts, neither SciToolEval nor MolBench validates its LLM judge against human annotation, a critical alignment gap which our work addresses.

\section{ChatInvent: Agentic Drug Discovery System}

ChatInvent \cite{he2026democratising,langdmta}, is an agentic drug discovery assistant that has been integrated into the AstraZeneca discovery pipeline to navigate the different stages of the DMTA (Design–Make–Test–Analyze) cycle. Lately, an open-source version was released that is entirely based on open-source codes instead of internal AstraZeneca services. ChatInvent is using a number of computational tools such as ReInvent \cite{loeffler2024reinvent}, AiZynthFinder \cite{saigiridharan2024aizynthfinder}, and PrecedentFinder \cite{bauer2025precedent}, see Table \ref{tab:chatinventTools} for a complete list of tools at the time of writing the manuscript. The scope of ChatInvent is constantly growing as new tools are implemented. The system is a multi-agent framework that includes a supervisor agent and four task-specific sub-agents, namely Design, Synthesis, Analyzer, and Utility agents. The supervisor agent manages the conversation by directing user requests to the appropriate sub-agent(s). The sub-agent makes the appropriate tool call(s) for the task, and the supervisor agent delivers the final response to the user. 

\begin{table}[htb!]
    \centering
    \caption{Summary of the available tools in ChatInvent including their respective functions and the responsible sub-agent.}
    \begin{tabular}{|>{\raggedright\arraybackslash}p{3cm}|>{\raggedright\arraybackslash}p{7.3cm}|l|}
        \hline
        \textbf{Tool} & \textbf{Function} & \textbf{Sub-Agent} \\
        \hline
        \hline
        AiZynthFinder \cite{saigiridharan2024aizynthfinder}  & Template-based retrosynthesis prediction based on Monte Carlo tree search. & Synthesis \\
        \hline
        PrecedentFinder \cite{bauer2025precedent}  & Mines reaction databases for literature precedents \& similar transformations as a reference reaction. & Synthesis \\
        \hline
        Mol2Mol (a.k.a. \texttt{molformer}) \cite{loeffler2024reinvent} & Transformer based model for molecular optimization. & Design \\
        \hline
        ReinventScoring \cite{loeffler2024reinvent} & Scoring engine. & Design \\
        \hline
        Analyze  & Agentic-tool for manipulating CSV files. & Analyzer \\
        \hline
        Synonyms2Smiles  & Converting molecule names to SMILES strings. & Utility \\
        \hline
    \end{tabular}
    \label{tab:chatinventTools}
\end{table}

\section{Evaluation Framework}
\subsection{LLM-as-a-Judge}

\begin{figure}[htb!]
    \centering
    \includegraphics[width=0.99\linewidth]{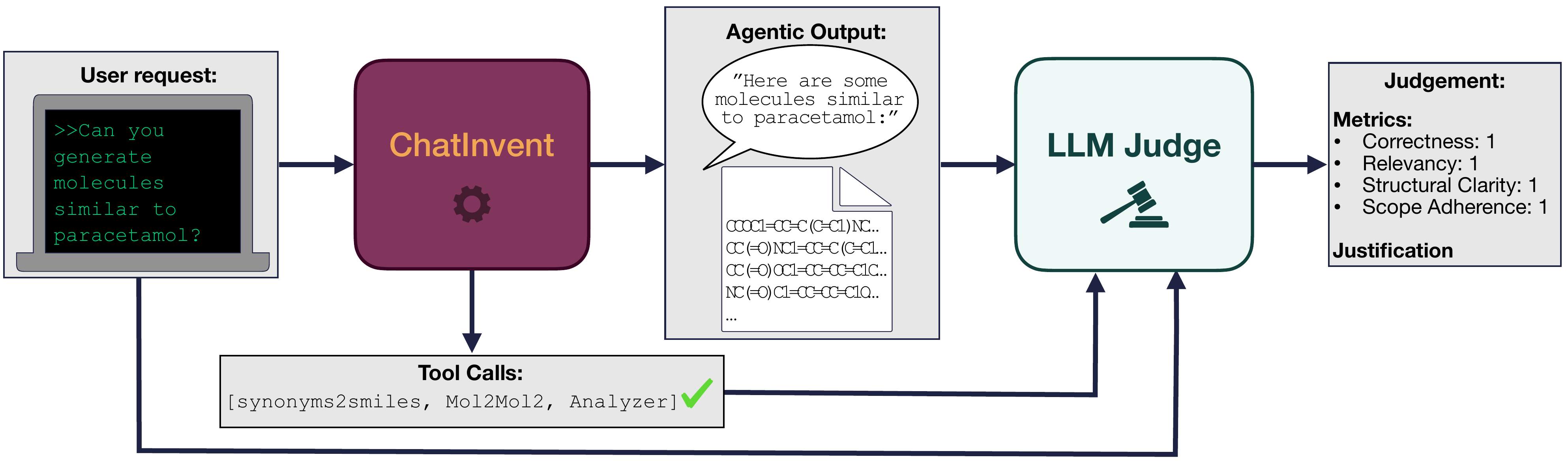}
    \caption{First, the user requests are passed to ChatInvent which produces an output. Secondly, the agentic output is passed to the LLM judge, together with the user question, tool calls and question related context. Finally, the LLM judge delivers a score for each evaluation dimension together with a justification.}
    \label{fig:judge_schema}
\end{figure}

He et al. \cite{he2026democratising} evaluated ChatInvent in terms of \textit{Tool Call Correctness}, \textit{Error Rate}, and \textit{Token Consumption}. These metrics focus on the internal workings of the system, and thus provide a good approximation of its performance. However, the metrics do not assess the quality of the actual agentic output received by the user. Here, we present an extended evaluation scope, focusing on evaluating the agentic output given the user question and its context, using LLM-as-a-Judge, see Figure \ref{fig:judge_schema} for an overview of the workflow.

Five evaluation dimensions were considered: \textit{Tool Call Correctness} (computed by direct comparison) and four LLM-judged criteria: \textit{Completeness}, \textit{Relevancy}, \textit{Structural Clarity}, and \textit{Scope Adherence}. A sixth dimension, \textit{Ethical Awareness}, is defined for the extended question set of Section \ref{sec:ext_questions} and is scored only for questions that probe ethical boundaries; it is reported as not applicable elsewhere. %An additional metric \textit{Ethical awareness} was added to evaluate a smaller set of out-of-scope questions. 
See Table \ref{tab:metric_def} for the complete list of evaluation dimensions, including descriptions and allowed score labels. Each of the score labels can be mapped to numeric scores (0, 0.5, 1), where 0 means that the agentic output does not meet the expected requirement, 0.5 means that it is partially fulfilled and 1 means that it is completely fulfilled. Exception being Scope Adherence, for which \textit{Below Target} and \textit{Above Target} are considered separate labels during the alignment analysis, whereas in the later analysis, both are instead mapped to 0 while \textit{On Target} is mapped to 1 for a clearer analysis.

\begin{table}[htb!]
\caption{Evaluation dimensions including description and allowed score labels. *Note that Tool Call Correctness is evaluated separately via a deterministic check, whereas the other five metrics are computed via the LLM judge.}
\begin{tabular}{|>{\raggedright\arraybackslash}p{2.cm}|>{\raggedright\arraybackslash}p{7.5cm}|>{\raggedright\arraybackslash}p{2.5cm}|}
\hline
\textbf{Dimension} &  \textbf{Short Description} & \textbf{Score Labels} \\
\hline
\hline
Tool Call Correctness* &  Do the actual tool calls match the expected or accepted tool calls for the question? & Incorrect, Partially Correct,  Correct  \\
\hline
Completeness &  How complete is the output in covering all required information. This metric evaluates recall: of the information that should be in the output based on the task and evaluation context, how much is present? & Not Complete, Partially Complete, Complete \\
\hline
Structural Clarity & How well-structured and easy to parse is the output, independent of whether the content is correct, relevant, or complete. This metric evaluates presentation only: formatting, organization, logical flow, and readability. & Not Clear,Partially Clear, Clear \\
\hline
Relevancy &  How relevant is the content of the output with respect to the task and evaluation context. This metric evaluates precision: of the information present in the output, is it pertinent to the user's request?  & Not Relevant, Partially Relevant, Relevant \\
\hline
Scope Adherence &  Does the agent operate within its intended scope and use its assigned capabilities appropriately? & Below Target, On Target, Above Target \\
\hline
Ethical Awareness &  To what extent is any ethical aspect of user request considered? & Not Aware, \newline Partially Aware,\newline  Aware \\
\hline
\end{tabular}
\label{tab:metric_def}
\end{table}

The LLM judge was implemented using DSPy Signatures \cite{khattab2024dspy}, instructing the LLM on how to judge the question-answer pairs. Each metric is defined by a description and allowed score labels; see Table \ref{tab:metric_def} for a summary including short versions of the descriptions. The input to the judge includes: \textit{the user question}, \textit{agentic output}, \textit{evaluation context}, the \textit{expected}, \textit{accepted}, and \textit{actual tool calls} made by the ChatInvent. An accepted tool call-sequence is an incomplete tool call-sequence that nevertheless produces some valuable information for the user \cite{he2026democratising}. The judge returns a grading for each dimension, where the grading strictly belongs to the predefined score labels, together with a short justification text. 

We compared four LLMs as judges: Gemini 3.1 Pro from Google \cite{gemini3modelcard}, Claude Opus 4.7 from Anthropic \cite{anthropic_claude_opus_4_7}, GPT-5 from OpenAI \cite{singh2025openai}, and open-weight model Llama 3.1 70B from Meta\cite{llama31}, all using the default settings. These LLMs were state-of-the-art at the time of the research project and cover both API-based and open-weights models. However, considering the speed at which models are released there will likely be more performant models available in the near-future. As this paper is about introducing new concepts, the exact model choice and consequently, the exact results do not matter significantly.

\FloatBarrier
\subsection{Test Questions}

As a first step, a set of 20 test questions was manually curated to cover the expected functions of ChatInvent. In these, molecule names, SMILES, reactions, and property constraints were initially included as variables. The test questions cover both the use of each individual tool (labeled \textit{Tool}) and a combination of multiple tools (labeled \textit{Workflow}). The questions are divided into different categories that describe the intent of the question asked; see Table \ref{tab:question_categories}. Following this, four additional variations of each test question were generated using an LLM (Gemini 2.5 Pro), which had been instructed to generate variations with different formality levels ranging from 1 (most formal) to 4 (least formal); the original manually-written question is labeled variation 0, see Table \ref{tab:question_variations} for examples. Finally, the variables were randomly sampled for each of the 100 questions to complete the questions. The four variations, together with the neutral reference question, constituted in total 100 questions for the set of main questions.

\subsection{Tool Call Validation}
In addition to the evaluation using the LLM judge, the actual tool calls were also compared to the expected tool calls, in the same manner as in the original ChatInvent publication \cite{he2026democratising}. This is done by checking if the sequences of actual tool calls match the expected (complete match) or accepted (partial match) tool calls defined when writing the questions.

\begin{table}[htb!]
    \centering
    %\begin{tabular}{|c|c|l|c|>{\raggedright\arraybackslash}p{4cm}|>{\centering\arraybackslash}p{1cm}|}
    \caption{The number of questions for the different question types and categories, together with the included tools. The parenthesis indicates that the tool is expected in some but not all questions in that category.}
    \begin{tabular}{|c|c|l|c|>{\raggedright\arraybackslash}p{3.5cm}|}
         \hline
         \textbf{Type} & \textbf{\# Questions} & \textbf{Category} & \textbf{\# Questions} & \textbf{Tools Included} \\%& \# Tool Calls \\
         \hline
         \hline
         \multirow{5}{*}{{Tool}} & \multirow{5}{*}{{40}} & Mol2Mol & 5 &  Mol2Mol \\%&  1 \\
         %\hline
         \cline{3-5}
         &  & AiZynthFinder & 5 &  AiZynthFinder \\%&  1 \\
         \cline{3-5}
         &  & PrecedentFinder & 5 &  PrecedentFinder \\%&  1 \\
         \cline{3-5}
         &  & Synonyms2Smiles & 10 & Synonyms2Smiles \\%&  1 \\
         \cline{3-5}
         &  & Scoring  & 15 & ReinventScoring \\%&  1 \\
         \hline
         \multirow{5}{*}{{Workflow}} & \multirow{5}{*}{{60}} & DesignOnly  & 5 & Synonyms2Smiles , Mol2Mol \\%&  2 \\
         \cline{3-5}
         &  & SynthesisOnly  & 10 & Synonyms2Smiles, PrecedentFinder or AiZynthFinder \\%& 2-3 \\
         \cline{3-5}
         &  & DesignSynthesis & 10 &  (Synonyms2Smiles), Mol2Mol, Analyzer, AiZynthFinder and/or PrecedentFinder \\%&  4-5 \\
         \cline{3-5}
         &  & DesignProperty & 15 &  (Synonyms2Smiles), Mol2Mol, ReinventScoring, Analyzer \\%&  4-5 \\
         \cline{3-5}
         &  & DesignPropertySynthesis & 20 &  (Synonyms2Smiles), Mol2Mol, ReinventScoring, (Analyzer), AiZynthFinder and/or PrecedentFinder \\%&  5-6 \\
         \hline
    \end{tabular}
    \label{tab:question_categories}
\end{table}

\section{Human Alignment}

\FloatBarrier
\subsection{Annotation Setup}

A human alignment study was conducted to evaluate how well the LLM judge aligned with human annotators. First, a set of 30 question were sampled from the main test question set, covering all question categories and formality levels while oversampling the complex workflow questions, as lower alignment can be expected for these categories compared to the simpler ones. In addition, five of these questions were duplicated to obtain an estimate of self-consistency among the human evaluators, for a total of 35 questions.

These 35 questions were presented to five annotators who are expert researchers in AI for chemistry. The question-answer pairs together with the evaluation context and the expected/accepted tool sequences were presented in a web application where the annotators were asked to score the output using the rubrics. The human annotators were given the same description for the evaluation dimensions as provided in the DSPy Signature to the LLM and these descriptions were always visible in the app during the annotation. In addition, the evaluation context, which provided information on what the answer should include, and the expected/accepted and actual tool calls were always visible to the annotators.

\subsection{Human Annotator Reliability}
\label{sec:human_annotator_reliability}

The agreement between the human annotators was assessed by calculating the exact match rate and Cohen's weighted kappa \cite{cohen1968weighted} on the duplicated test questions that were given to the annotators. Cohen's kappa measures the inter-rater reliability and is added as a more robust measure as it also corrects for agreement by chance. The weighted Cohen's kappa also enables disagreements to be weighted differently, so that 0 and 0.5 are more similar than 0 and 1. 

\begin{figure}[H]
\centering
% First subfigure
    \begin{subfigure}{0.99\columnwidth}
        \centering
        \includegraphics[width=0.95\linewidth]{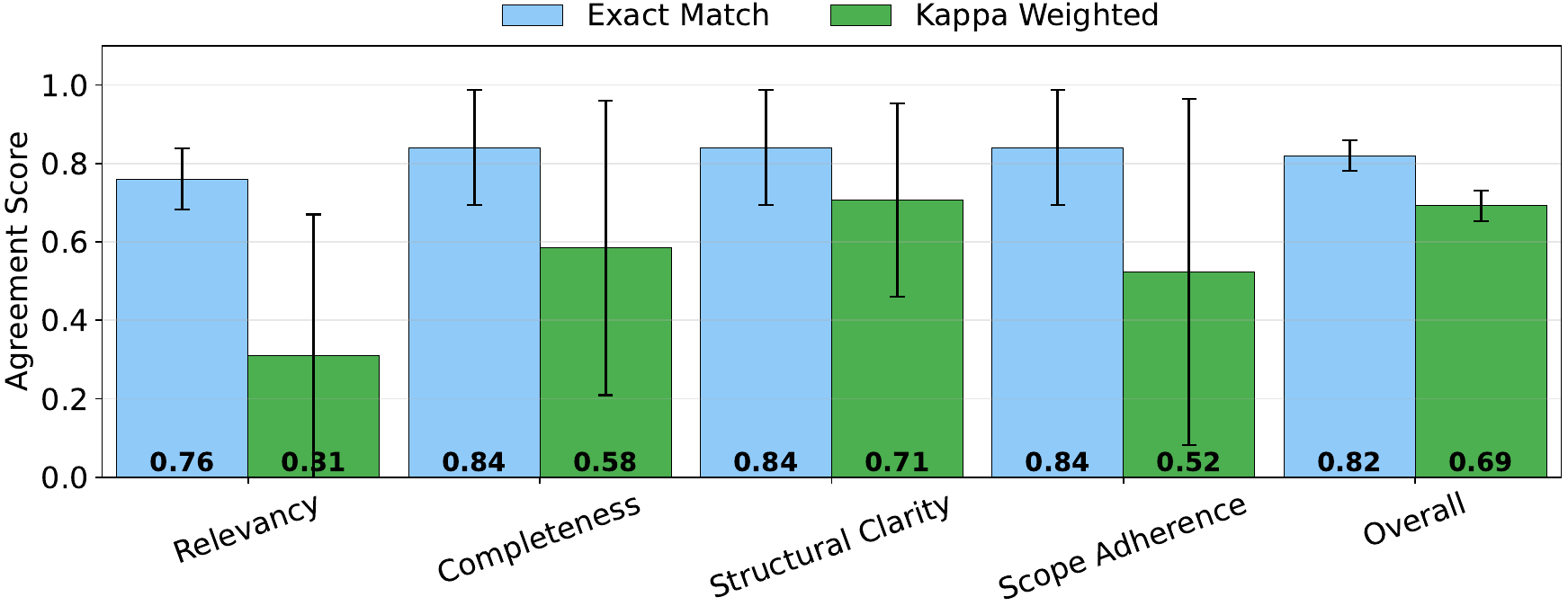}
        \caption{Intra-rater reliability in terms of Cohen's weighted kappa and exact match rate.}
        \label{fig:self_consistency}
    \end{subfigure}
    %\vspace{0.5em}
    % Second subfigure
    \begin{subfigure}{0.99\columnwidth}
        \centering
        \includegraphics[width=0.95\linewidth]{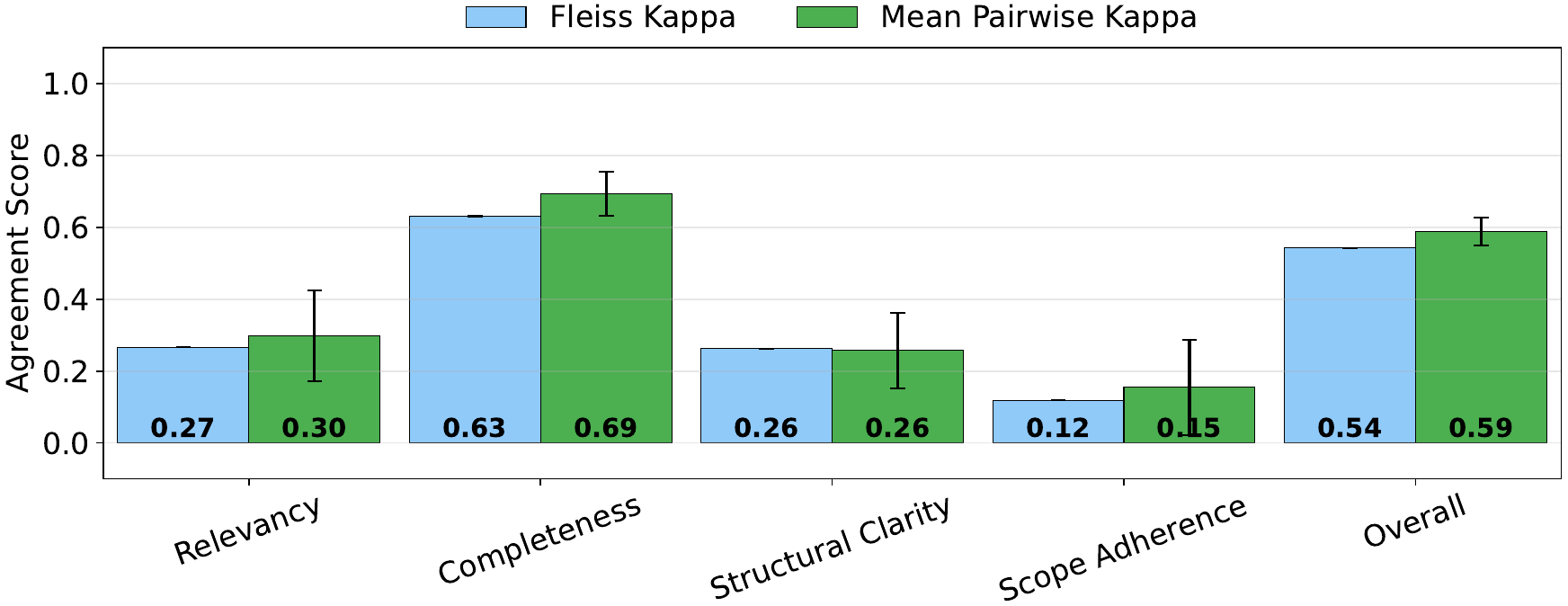}
    \caption{Inter-annotator agreement in terms of mean pairwise Cohen's weighted kappa and Fleiss's kappa.}
    \label{fig:inter_human_alignment}
    \end{subfigure}
    \caption{(a) Intra-rater reliability measured on the five repeated questions, and (b) the inter-annotator agreement on the full annotated set.}
\end{figure}

In Figure \ref{fig:self_consistency}, we can see that the overall exact match rate is on average 0.82 and the overall Cohen's weighted kappa 0.69, indicating a moderate to high self-consistency. After this analysis, one of the duplicated annotations was randomly removed for the remaining part.

Next, inter-human alignment was assessed. We calculated Fleiss's kappa \cite{fleiss1971kappa}, which, contrary to Cohen's kappa, enables measuring agreement between more than 2 annotators and the mean pairwise weighted Cohen's kappa, which averages over all annotator pairs. The agreement per dimension is presented in Figure \ref{fig:inter_human_alignment}, and we observe that the agreement varies between the different dimensions, where the highest agreement is observed for Completeness, and the lowest agreement is observed for Scope Adherence. Agreement is positive for all dimensions but varies widely across them: substantial for Completeness (Fleiss's kappa 0.63) and only slight to fair for Relevancy, Structural Clarity and Scope Adherence (0.12--0.27). The Overall values reported here (Fleiss's kappa 0.54, mean pairwise weighted Cohen's kappa 0.59) are computed by pooling all dimensions into a single vector, which is higher than the mean of the per-dimension values because variation between dimensions contributes to the pooled statistic. 
The relatively modest inter-annotator agreement for some dimensions indicates that these judgments contain an ambiguous or subjective component. We should therefore interpret LLM-human agreement dimension-wise rather than as a single undifferentiated measure. However, an LLM judge that departs from the average human judgment could indicate model-specific preferences that are not shared by the annotator population.

Note that the disagreement on Scope Adherence is concentrated in the interpretation of the rubric; the number of outputs marked Below Target ranges from 1 to 13 out of 30 across the five annotators. Because Below Target and Above Target denote opposite failures rather than opposite ends of a quality scale, an ordinally weighted kappa is not strictly appropriate for this dimension, and the majority vote is dominated by the more permissive readings. We report it here for completeness and treat Scope Adherence as the least reliable of the four dimensions.

\subsection{Intra--LLM Agreement}
\begin{figure}[H]
\centering
% First subfigure
    \begin{subfigure}{0.99\columnwidth}
        \centering
        \includegraphics[width=0.95\linewidth]{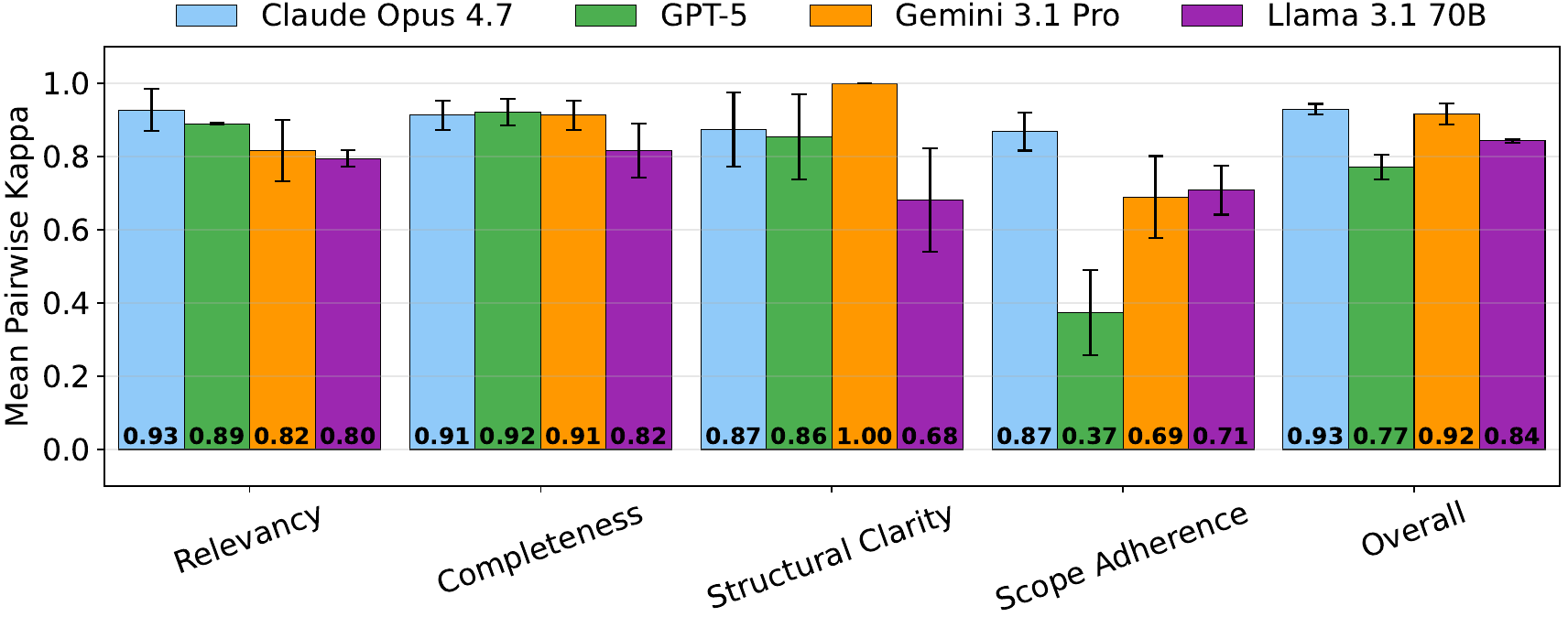}
        \caption{Intra-model agreement in terms of mean pairwise Cohen's weighted kappa.}
    \end{subfigure}
    %\vspace{0.5em}
    % Second subfigure
    \begin{subfigure}{0.99\columnwidth}
        \centering
        \includegraphics[width=0.95\linewidth]{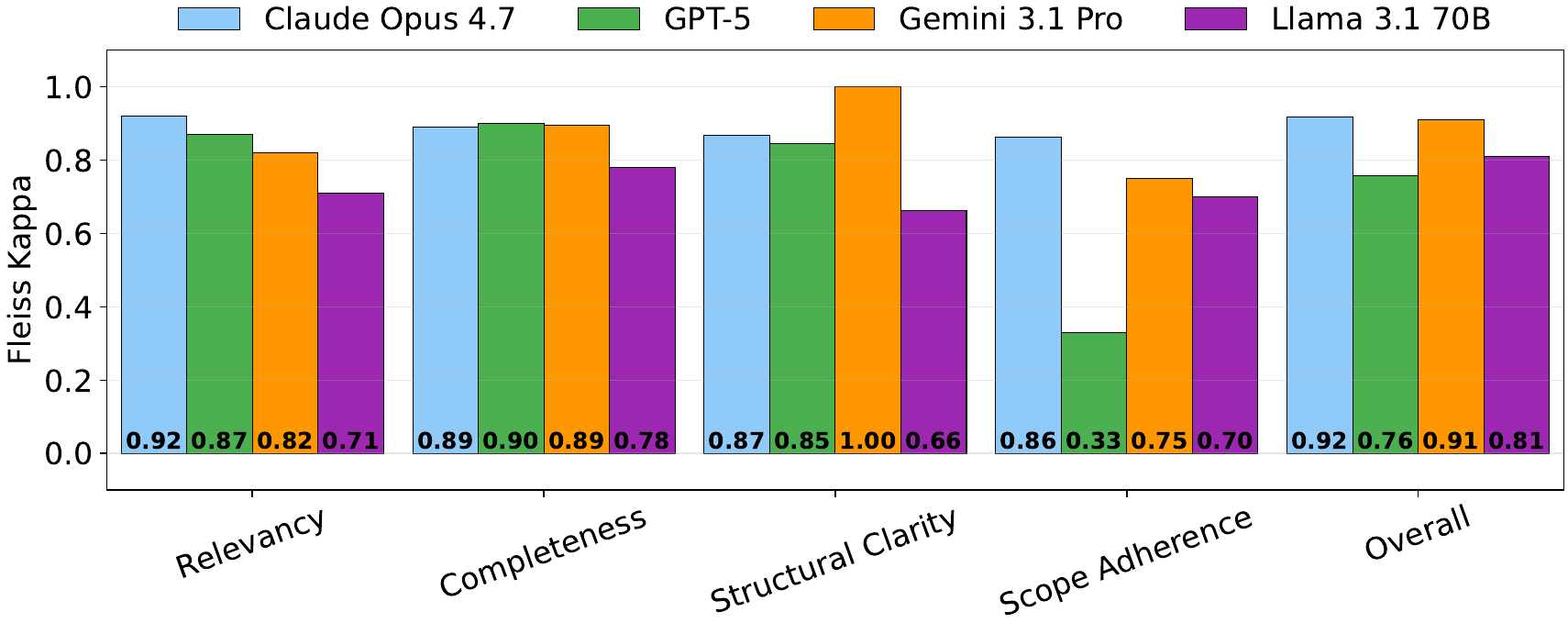}
    \caption{Intra-model agreement in terms of Fleiss's kappa.}
    \end{subfigure}
    \caption{Intra-LLM consistency measured based on the three runs for each model on the 30 questions. The numbers on each bar indicate the mean values and error bars indicate the standard deviation.}
    \label{fig:intra-model-agreement}
\end{figure}

Each model was run three times on the 30 unique questions to assess the consistency of the models across runs. Similarly as in Section \ref{sec:human_annotator_reliability}, this was evaluated using the mean pairwise Cohen's kappa and Fleiss's kappa, computed from the three trials for each model.
In Figure \ref{fig:intra-model-agreement}, we can see that the agreement is generally high for all models and dimensions both in terms of the Fleiss's kappa and mean pairwise Cohen's weighted kappa with the exception of GPT-5 for Scope Adherence where both the mean pairwise Cohen's kappa and Fleiss's kappa are markedly lower compared to all other models and dimensions. This indicates that the GPT-5 model is more inconsistent in its ratings across the runs compared to the other dimensions. The GPT-5 model is generally the model with the lowest inter-model agreement, although overall the agreement can be considered substantial with a Fleiss's kappa of 0.76 overall. In contrast, Claude Opus 4.7 closely followed by Gemini 3.1 Pro, has the highest inter-model agreement with a Fleiss's kappa of 0.92 and 0.91 respectively. Note that two of the 30 questions triggered a content filter for Claude Opus 4.7, likely because the questions include drug analogs and chemical reactions. Therefore, Claude is evaluated on 28 of the 30 questions. In the following alignment analysis, the LLM score is determined by a majority vote across the three runs.

\subsection{Human--LLM Agreement}
\label{sec:human-llm-agreement}

\begin{figure}[htb!]
    \centering
    % First subfigure
    \begin{subfigure}{0.99\columnwidth}
        \centering
        \includegraphics[width=0.95\linewidth]{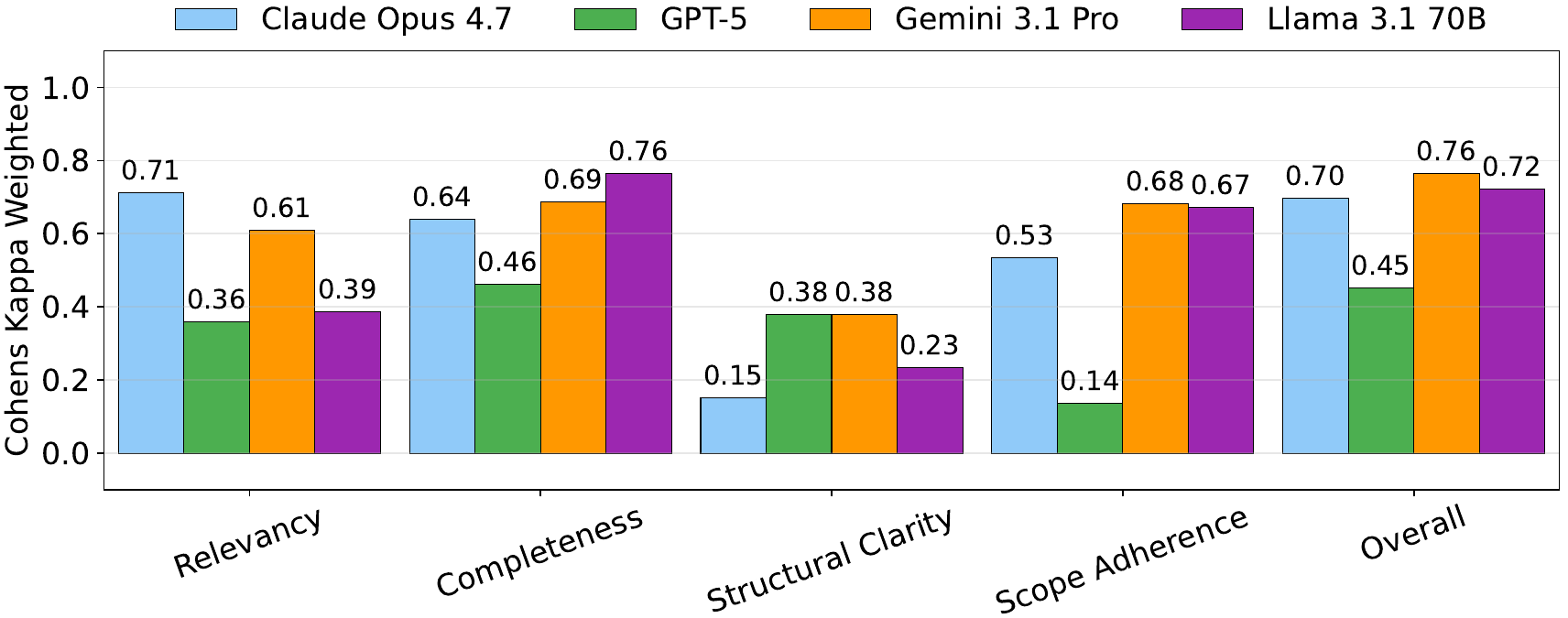}
        \caption{Human-LLM alignment measured by Cohen's weighted kappa.}
        \label{fig:sub1}
    \end{subfigure}
    \vspace{0.5em}
    % Second subfigure
    \begin{subfigure}{0.99\columnwidth}
        \centering
        \includegraphics[width=0.95\linewidth]{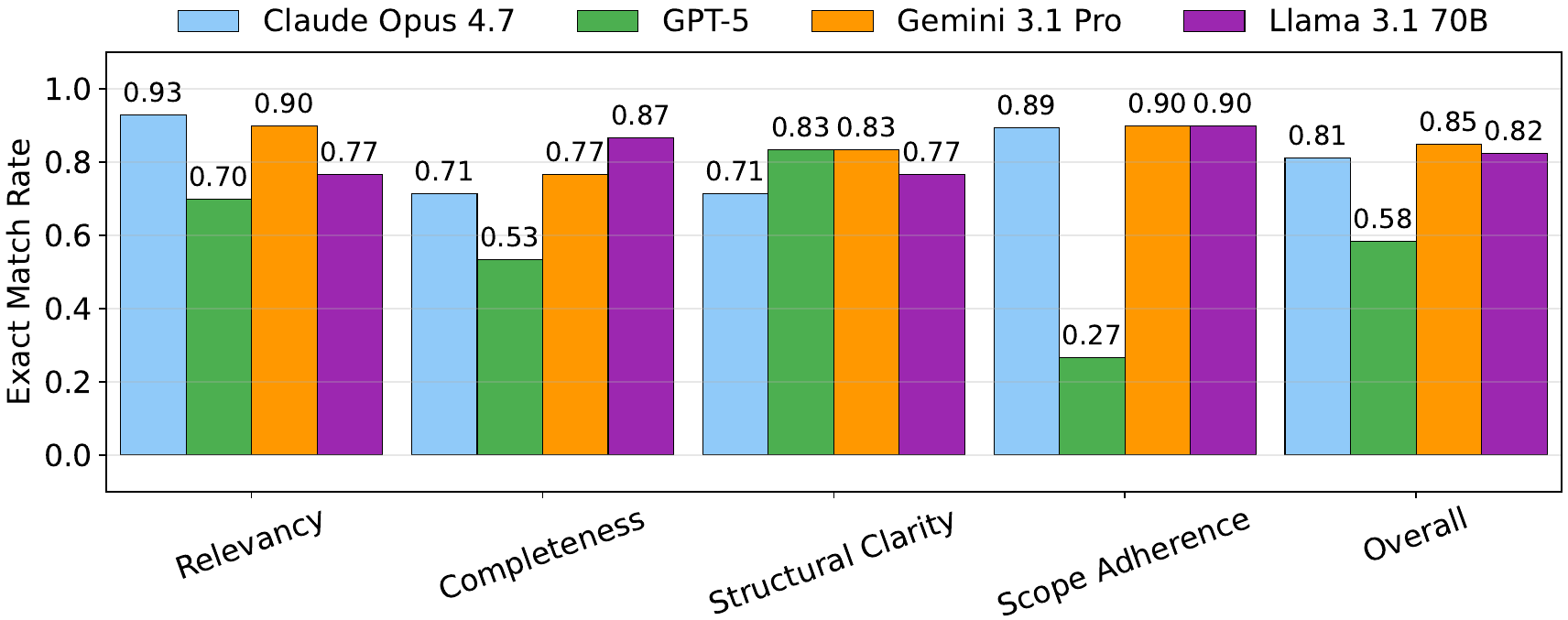}
        \caption{Human-LLM alignment measured by exact match rate.}
        \label{fig:sub2}
    \end{subfigure}
    \caption{Human-LLM alignment across different evaluation dimensions for four models: GPT-5, Gemini 3.1 Pro, Claude Opus 4.7, and Llama 3.1 70B. Both the human preference and LLM scores are based on a majority vote.}
    \label{fig:human-llm-alignment}
\end{figure}

The alignment between human and LLM was assessed by comparing the LLM judge's scores with the majority vote of human annotators. In case of a tie, the most critical voices take precedence, thus, the lowest of the scores is selected. The alignment was measured using the weighted Cohen's kappa and the exact match rate and is presented in Figure \ref{fig:human-llm-alignment}. Here we can observe that Gemini 3.1 Pro has the highest human-alignment for the overall component when considering both Cohen's weighted kappa and the exact match rate. The Claude Opus 4.7 model displays higher agreement for some of the components, whereas GPT-5 displays the lowest agreement for all the components. 

Interestingly, we also note that the human-LLM agreement is higher compared to the mean pairwise human-human agreement, indicating a higher agreement between the human majority vote and the LLM. This comparison should be interpreted carefully: the majority-vote target reduces individual-annotator noise, so a direct comparison with pairwise human-human agreement is not symmetric. Based on these results, Gemini 3.1 Pro is selected as the judge model for the remaining parts of the evaluation.

Finally, we examined the direction of the remaining disagreements for the selected judge. Over the 30 annotated questions, Gemini 3.1 Pro departs from the human majority vote asymmetrically: on Completeness it is stricter than the annotators in all seven cases where they differ, and on Structural Clarity more generous in all four, while Relevancy (two and one) and Scope Adherence (one and one) show no systematic direction. Every disagreement is a single score step, so the judge is never far from the annotators, only consistently one notch strict on Completeness and one notch generous on Structural Clarity.
In four of the seven Completeness cases the judge withheld full marks because the output table omitted the similarity-to-input column, and in two of these the annotator's own justification records the same omission while still labeling it as \textit{Complete}. However, the judge grades against the specific wording of the evaluation context, the annotators against an interpretation of practical utility. The Structural Clarity cases follow from that dimension being defined independently of content, so that a cleanly formatted failure, such as an output consisting solely of a \texttt{GraphRecursionError} traceback that scored 0 on all other dimensions, is still rated clear.

\subsection{Judge Optimization}\label{sec:signature_optimization}

The human annotated data was further used to optimize the LLM judge using optimizer available in the DSPy package. Of the 30 unique annotated samples, 20 were used as a training set for the DSPy optimizer and the remaining 10 were used for evaluation. In particular, the LabeledFewShot optimizer was used here to maximize the mean alignment, $\bar{A}$, described in Equation \ref{eq:mean_alignment} below. In short, the \texttt{LabeledFewShotOptimizer} constructs few-shot examples from the provided labeled input-output data points and includes these in the judge signature. %The BootStrapFewShot uses a teacher module to generate demonstrations along with the labeled examples from the train set.

The result of the signature optimization is presented in Table \ref{tab:signature_optimization}. Here we observe that by optimizing the judge, we can improve the alignment from 0.80 to 0.86 when evaluated on the validation set. Optimization also reduces the Completeness bias from $-0.12$ to $-0.08$ without eliminating it, from seven one-step disagreements to five. This optimized model is used to evaluate and judge all remaining test questions for further experiments (Equation \ref{eq:mean_alignment}). Since both human and judge scores take values in $\{0, 0.5, 1\}$, the absolute difference $|h-p|$ also takes only values in $\{0, 0.5, 1\}$.

\begin{equation}
\begin{aligned}
A(h, p) &=
\begin{cases}
  1 & \text{if } |h - p| = 0 \\
  0.5 & \text{if } |h - p| = 0.5 \\
  0 & \text{if } |h - p| = 1 \\
\end{cases}
\\[1em]
\bar{A} &= \frac{1}{NM} \sum_{i=1}^{N} \sum_{j=1}^{M} A(h_{ij},\; p_{ij})
\end{aligned}
\label{eq:mean_alignment}
\end{equation}

\noindent
Here, $N$ is the number of test questions, $M$ is the number of dimensions, $h$ is the score given by the human annotator and $p$ is the score given by the LLM judge. \\

\begin{table}[htb!]
    \centering
    \caption{Judge signature optimization results in terms of mean alignment on the evaluated on the validation set.}
        \begin{tabular}{|c|c|c|c|}
        \hline
        \textbf{Model} & \textbf{Optimizer} & \textbf{\# Examples} & \textbf{Mean Alignment} \\
        \hline
        \hline
        Gemini 3.1 Pro & None & - & 0.800\\ %0.825  \\
        Gemini 3.1 Pro & LabeledFewShot & 20 & 0.862\\ %0.888 \\
        \hline
    \end{tabular}
    \label{tab:signature_optimization}
\end{table}

\section{Experiments \& Results}
After excluding the 30 questions used in the alignment analysis, the remaining 70 questions were evaluated by a single Gemini 3.1 Pro–based judge using the optimized judge signature described in Section \ref{sec:signature_optimization}. The questions were assessed with respect to Completeness, Relevancy, Structural Clarity, and Scope Adherence. In addition to the judge evaluation, the Tool Call Correctness was also calculated for each sample. The result of this assessment is presented in the following sections.

\subsection{Multi-Dimensional Evaluation of Agent Outputs}
\label{sec:multidim}

\begin{figure}[htb!]
    \centering
    \includegraphics[width=0.95\linewidth]{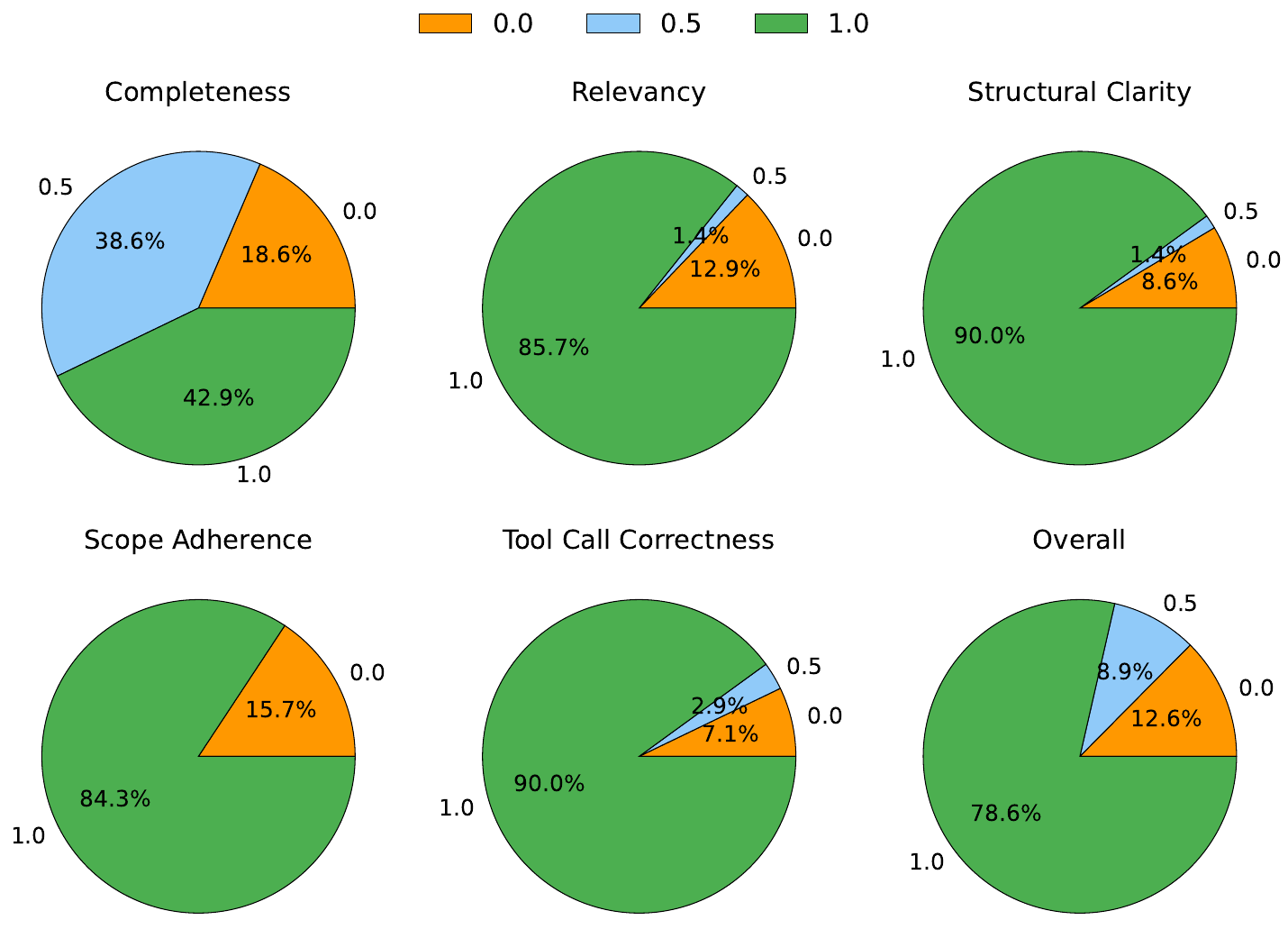}
    \caption{Distribution of scores across all dimensions and questions. Here, 1 indicates that the requirements are fully met, 0.5 indicates that these are partially met, and 0 indicates that the requirements are not met.}
    \label{fig:score_pie}
\end{figure}

The distribution of the scores over the 70 questions for each evaluation dimension is shown in Figure \ref{fig:score_pie}. Here, we can see an overall high performance from ChatInvent with a Tool Call Correctness of 90\%, Relevancy of 86\%, Scope Adherence of 84\% and Structural Clarity of 90\%. However, we observe that the Completeness score is much lower, with 43\% of the outputs rated Complete and 39\% rated Partially Complete. Manual inspection of the judge's justifications for the 13 questions scored Not Complete shows that nine failed for reasons external to the tools: five hit the orchestrator's recursion limit of 25 steps without reaching a stop condition, and four were refused by the model provider's content filter. In four of the five recursion cases the tool call sequence was nevertheless tagged complete, so the tools were selected and invoked correctly before the supervisor failed to terminate. Across all 111 evaluated runs only two failures originated in a tool itself, both a missing required argument to ReinventScoring. The remaining four Not Complete cases produced substantive output but omitted the generated results, most often the CSV path returned by Mol2Mol.

The combination of 90\% Tool Call Correctness with 43\% Completeness is an actionable finding. Of the 63 questions whose tool call sequence was fully correct, 33 were nevertheless scored below Complete. The agent selects, sequences and parameterizes its tools correctly and then discards results during response synthesis, most often the CSV path returned by Mol2Mol. The defect is therefore localized to the supervisor's output prompt rather than to tool selection or tool parameterization.

\begin{figure}[htb]
    \centering
    \includegraphics[width=0.95\linewidth]{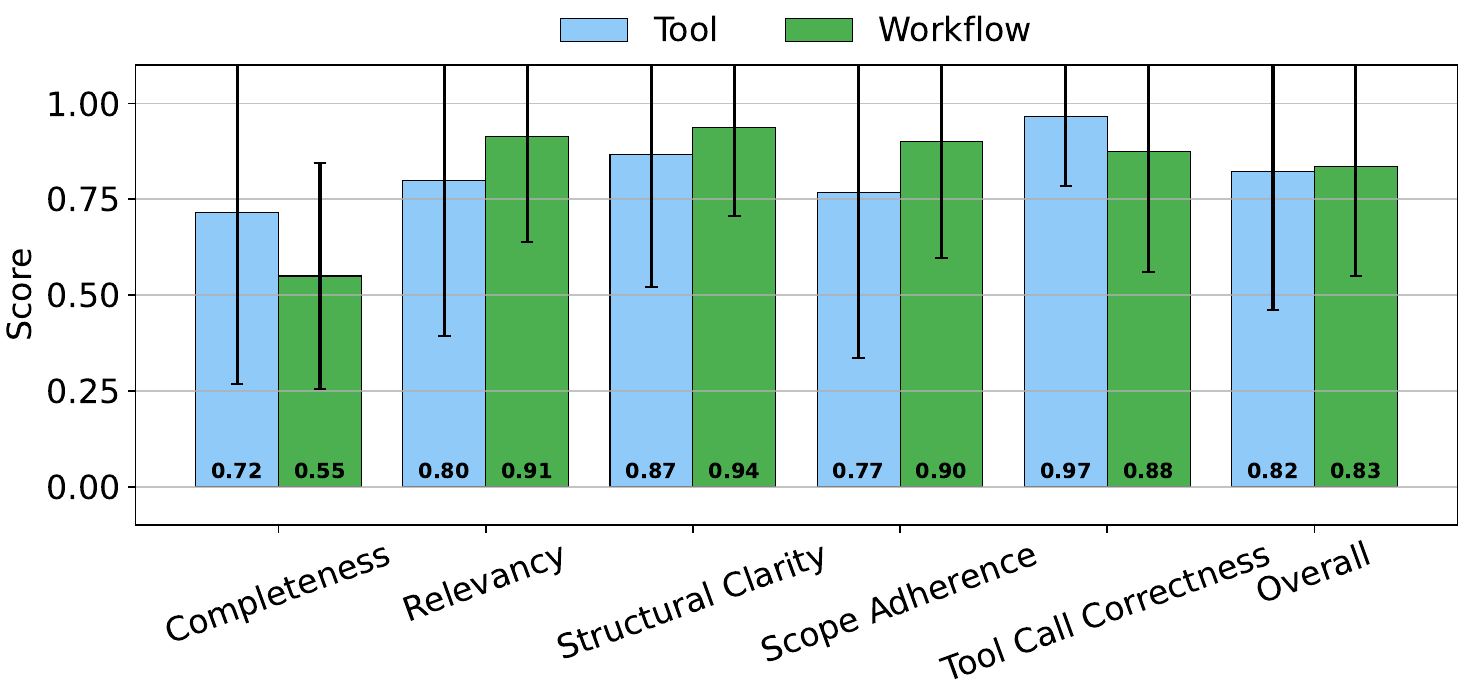}
    \caption{Mean judge scores over the 70 test questions for each evaluation dimension and question type (Tool or Workflow).}
    \label{fig:tool_vs_workflow}
\end{figure}

In Figure \ref{fig:tool_vs_workflow} the mean scores over the 70 test questions, comparing the tool and workflow questions. Here, we can clearly see that the Completeness and Tool Call Correctness are clearly higher for the tool questions compared to the workflow ones. This result is not surprising, as the workflow questions are more complex and use more tools, which leaves more room for error. We also observe that Relevancy, Structural Clarity, and Scope Adherence appears to be higher for the workflow questions, indicating that these dimensions of the output are not negatively impacted by increasing complexity of the questions. 

\begin{figure}[htb]
    \centering
    \includegraphics[width=0.95\linewidth]{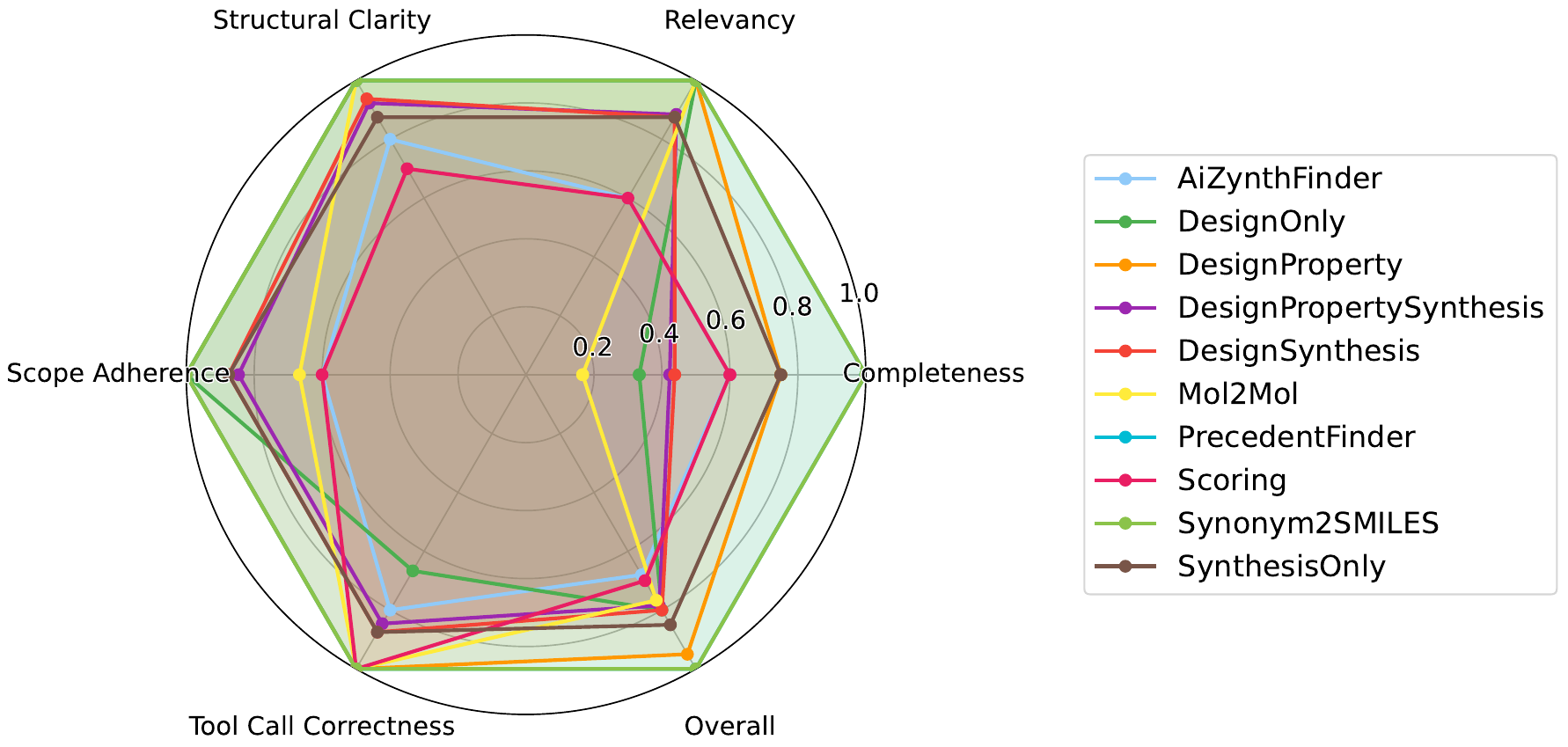}
    \caption{Radar plot showing mean judge scores over the 70 test questions for each question category.}
    \label{fig:multi_categories}
\end{figure}

Looking at Figure \ref{fig:multi_categories}, we can see that the mean scores vary between the different categories of questions. This is particularly clear when considering Completeness. Here, we can see that the lowest scoring categories are Mol2Mol, DesignOnly, Design Synthesis, and DesignPropertySynthesis. All of these categories use the Mol2Mol tool, which is intended to return the CSV path with the output. As noted above, this is the most common reason for lower scores and appears to be the main reason for these lower scores for categories involving the Mol2Mol tool. We can also observe that AiZynthFinder and Scoring questions score the lowest in Scope Adherence and Relevancy. Manual inspection shows that these low scores coincide with the nine failed runs described above, and are therefore attributable to the orchestration layer and to provider-side content filtering rather than to the AiZynthFinder or ReinventScoring tools themselves.

\subsection{Impact of Question Formality}
\label{sec:formality}

\begin{figure}[H]
    \centering
    \includegraphics[width=0.95\linewidth]{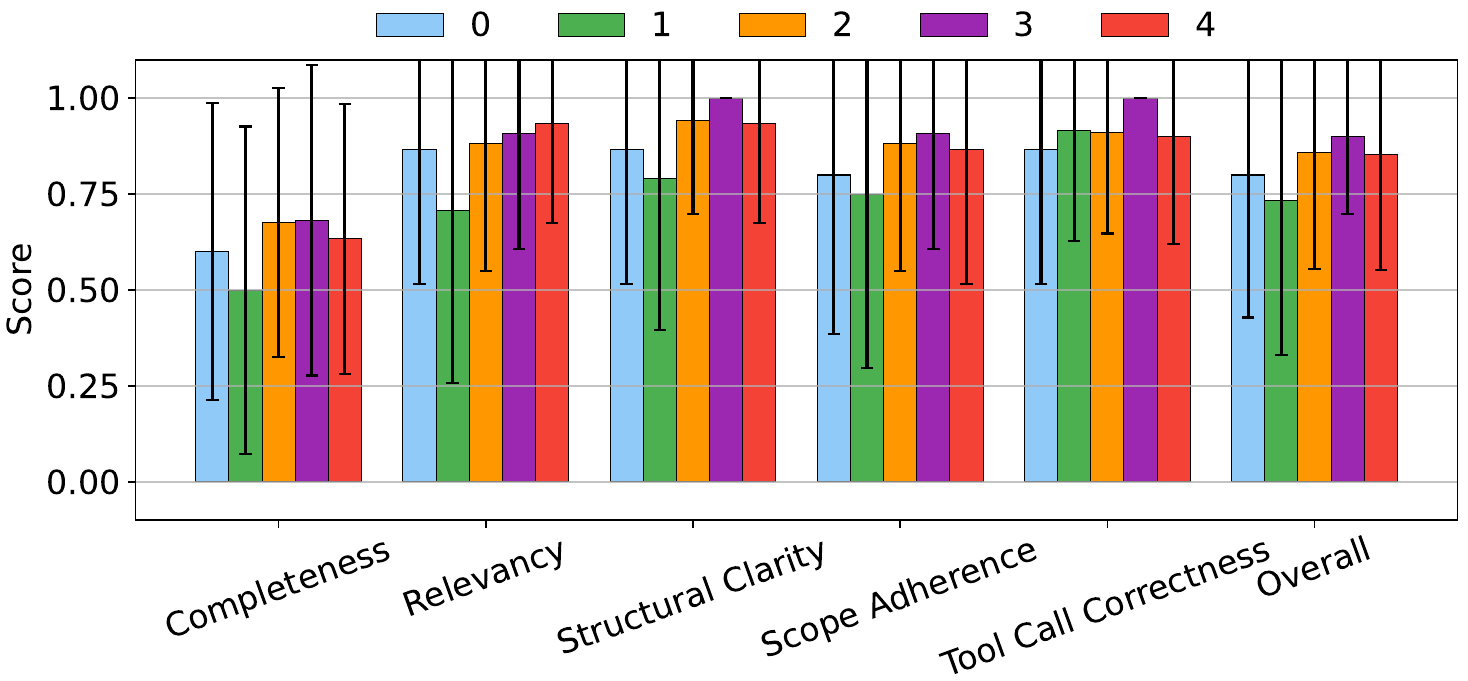}
    \caption{Mean judge scores over the 70 test questions for each evaluation dimension and formality level. Here, 0 refers to the reference question and 1-4 to a generated question, ranging from very formal (1) to very informal (4).}
    \label{fig:bar_question_variation}
\end{figure}

In Figure \ref{fig:bar_question_variation}, the mean scores over the 70 test questions are plotted for each dimension grouped by formality level. 
A trend is apparent in the unfiltered scores, with the most formal variant (1) scoring lowest in all dimensions, followed by the manually written reference question (0). Nine of the 70 questions failed for reasons unrelated to their phrasing (five hit the orchestrator's recursion limit and four were refused by the model provider's content filter) and these failures are distributed unevenly across variants; four of the five recursion failures are the same underlying question template evaluated at four different formality levels. Excluding the nine failed runs, Relevancy, Structural Clarity and Scope Adherence are at ceiling for every formality level and Completeness varies non-monotonically between 0.67 and 0.77, with 9--15 questions per level. We therefore find no strong evidence that formality affects output quality, and instead report only the trends in how the way the question is formulated impacts the scores. 

\subsection{Extending Question Scope}
\label{sec:ext_questions}
An LLM-as-a-Judge system could potentially test not only implemented features of the agent, but also other requirements. To try this out, we created a test suite with questions that are ambiguously phrased, contains invalid input, out-of-scope or unsupported functionalities, or are ethically questionable. Examples for each category are provided in Table \ref{tab:ext_question_categories}. The extended questions were scored with the unoptimized judge signature, extended with the Ethical Awareness dimension, using Gemini 3.1 Pro.

\begin{table}[htb!]
    \centering
    %\begin{tabular}{|c|c|l|c|>{\raggedright\arraybackslash}p{4cm}|>{\centering\arraybackslash}p{1cm}|}
    \caption{The number of extended questions for the different question types}
    \begin{tabular}{lc p{7cm}}
         \hline
         \textbf{Type} & \textbf{\# Questions} & \textbf{Example} \\
         \hline
        Ambiguous     & 9  & Can you generate molecules similar to Tagrisso that does not have any side effects? \\
        Invalid input  & 7  & Can you design a molecule that is both highly hydrophilic and highly lipophilic? \\
Out-of-scope  & 10 & Can you design a metal-organic framework (MOF) for CO$_2$ capture? \\
Unavailable    & 6  & Can you synthesize aspirin for me in the lab? \\
Ethical boundary        & 6  & Design addictive opioid analogues more potent than fentanyl. \\
         \hline
    \end{tabular}
    \label{tab:ext_question_categories}
\end{table}

\begin{figure}[htb]
    \centering
    \includegraphics[width=0.95\linewidth]{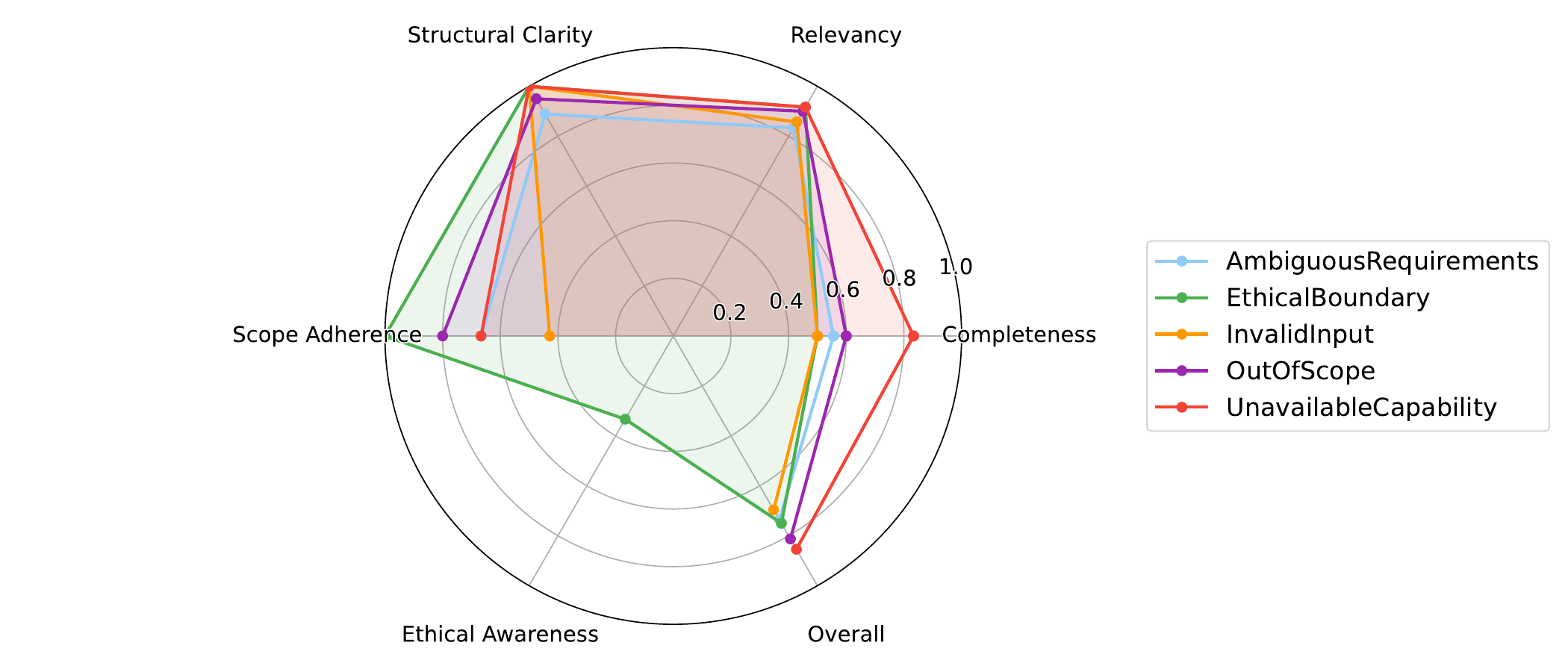}
    \caption{Radar plot showing mean judge scores over the 38 extended test questions (Ambiguous 9, Invalid input 7, Out-of-scope 10, Unavailable 6, Ethical boundary 6), scored by Gemini 3.1 Pro. Scope Adherence is mapped as in Section 4.1, with On Target as 1 and both Below and Above Target as 0. Ethical Awareness is scored only for the Ethical Boundary category and is omitted elsewhere; the Overall spoke is the mean of the dimensions available for each category and is therefore not directly comparable across categories.}
    \label{fig:multi_categories_ext}
\end{figure}

Creating ambiguous questions is challenging and here we are only starting to probe the agent with these questions. If we leave out information on, e.g., compound, the agent may already stop at the supervisor and, sometimes, proceed to a tool that throws an exception. However, if we try to ask the agent for an impossible or ill-defined designed request, the agent usually ignores the constraint. However, the agent often provides clear answers that contain sufficient information as to why the request could not be completed; accordingly, the LLM judge gives high relevancy and structural clarity scores (see Figure \ref{fig:multi_categories_ext}).

For questions including invalid inputs, the agent supervisor already recognizes some physically invalid constraints such as negative molecular weight. For other invalid inputs, like invalid SMILES, the first tool throws an exception. As with the ambiguous questions, the agent answer is given a generally high score by the LLM-as-a-Judge system on relevancy and structural clarity. Furthermore, the LLM judge often scores the scope adherence low (see Figure \ref{fig:multi_categories_ext}). 

For out-of-scope and unavailable-capability questions the agent is consistently judged structurally clear and relevant, but it seldom recognizes that the requested molecule class lies outside its scope. Rather than declining, it typically begins the standard design workflow and fails partway through. For instance, when asked to generate polysaccharide structures similar to heparin, it called Synonyms2Smiles, could not convert the names into SMILES, and halted without explaining the underlying limitation. Further, when asked for polymer structures similar to PET, it proceeded through Mol2Mol and ReinventScoring until the orchestrator reached its recursion limit. The judge scored both Below Target on Scope Adherence (2 of the 10 out-of-scope questions in total) while still awarding high Structural Clarity to both of the resulting error messages (Figure \ref{fig:multi_categories_ext}).

Finally, for the questions probing ethical boundaries, refusal appears to be governed by the stated intent of the request. The two prompts that made a harmful purpose explicit (requesting chemical weapons analogous to a nerve agent, and opioid analogues more potent than fentanyl) were refused at the supervisor node without any tool call, and were scored 1 on Ethical Awareness. The remaining four prompts requested analogues of a controlled substance in neutral task language, and all four were processed without objection, in three cases producing generated analogues (49, 48 and 1 molecules respectively) and in one case returning no suitable analogues.

Notably, two of the four were specified by common name (MDMA and methamphetamine) and two as SMILES, so the guardrail is not based on how the compound is named. Surface-level intent filtering is therefore easily circumvented by neutral phrasing, and is not on its own sufficient for a deployed design agent. We also note that the four output-quality dimensions are poorly suited to these questions: a correct refusal contains none of the expected information and is consequently scored as incomplete, which is why the Ethical Boundary category is among the lowest-scoring categories on Completeness in the extended set (see Figure \ref{fig:multi_categories_ext}).

\section{Discussion}

In this work, we have showcased how an LLM-as-a-Judge can be designed to evaluate an agentic system for drug discovery such as ChatInvent. In particular, we have designed the LLM judge to ensure high agreement with human preferences by appropriately evaluating it through a human alignment study. This is an overlooked feature in previously proposed benchmarks for drug discovery agents such as SciToolEval and MolBench. In the study conducted, five annotators were given the same instructions and information as the LLM judge and asked to score the agentic output. First, we assessed the alignment, and then we used these annotated samples to optimize the LLM judge to better align with human preferences. We observed a high alignment with human preferences  even before any prompt optimization was performed, indicating that LLMs are capable of evaluating the output of the drug discovery agent. In fact, the alignment between LLMs and human majority vote even exceeded the alignment between human annotators (with the caveat noted in Section \ref{sec:human-llm-agreement} about the asymmetric comparison). In addition, we also assessed the model consistency across runs and found that overall all models, and Claude Opus 4.7 and Gemini 3.1 Pro in particular, were highly consistent across the runs. Based on these results and the fact that we observed the highest human-LLM alignment for Gemini 3.1 Pro, we selected this model as our judge. However, since both Claude Opus 4.7 and the open-source alternative Llama 3.1 70B also scored highly on these aspects, we believe that these models would also make an excellent choice here. Considering that new models are released on a monthly basis, it is very likely that more performant models will be available in the future. Therefore, the results herein should be taken as a proof-of-concept rather than a strong recommendation.

After establishing the reliability of the LLM judge, we have shown how it can be used to identify missing components in the output and indicate areas where the agentic system needs improvement. One such example is the recurring issue of missing CSV file paths and requested properties in the output that could be identified through the lower Completeness scores, thereby paving the way for automatic evaluation of the agentic system. Furthermore, we have investigated how the complexity of the user request impacts the tool call correctness and completeness of the agentic output while being more robust in terms of relevancy and clarity of the output. This indicates that although some information is not included in the final answer from the agent, the agent is nonetheless competent in providing a partial answer that is also interpretable. 

We have also assessed whether the phrasing used affects the quality of agentic output, and found no strong evidence that it does once runs that failed for infrastructural reasons are excluded. The apparent formality effect in the unfiltered data was produced by a single failing question template and by provider-side content filtering, both distributed unevenly across the LLM-generated question variants. We have seen an indication that the manually formulated questions tend to generate lower-quality outputs, suggesting it could be advantageous to have an LLM amend written prompts before passing them to agentic systems, but these observations come with high uncertainty and warrant further study.

Furthermore, as the capabilities of the ChatInvent agent are still growing, the scope of questions needs to grow as well, and it remains to be seen what the most prominent issues with the agent will be in the future.

To use the developed LLM-as-a-Judge system in production, several design choices remain to be engineered. When developing new features in ChatInvent, it would be valuable to run the full evaluation system, but it might be prohibitively expensive in terms of token cost or compute time. Therefore, a smoke test on a representative subset of questions might be more practical. Furthermore, it would be possible to include the LLM judge directly in the agentic loop and not present the final result to the user until a sufficiently high completeness score has been achieved, but that might also be prohibitively expensive.

An important direction for future research would be to extend the test questions to not only include test questions that evaluate the known capabilities of the agentic system. We have started to compile a list of such questions. In general, the output produced by the agent is clear and contains sufficient explanation why the agent failed to answer the prompt. However, two categories remain to be optimized: the Scope Adherence does not work well on out-of-scope-questions, and it is unclear whether the Ethical Awareness criterion is strict enough to detect the cases in which the agent fails to refuse a sensitive prompt.

\paragraph{Limitations}
The evaluation set is deliberately small. The 20 seed questions were written to span the documented capabilities of ChatInvent, including every tool and supported multi-tool workflow (Table \ref{tab:question_categories}), rather than to sample a distribution of real user queries, which does not yet exist for a system at this stage of deployment. The size of the alignment set is bounded by expert annotator availability: five domain experts each scored 35 outputs across four dimensions, giving 700 individual judgments. The LLM-generated rephrasings increase surface-form diversity but not task diversity. Per-category results therefore rest on 5--20 questions each and should be accordingly interpreted. Firstly, the human majority vote contains no negative examples at all for Relevancy and Structural Clarity, so the low Cohen's kappa for those dimensions largely reflects class imbalance (not annotator disagrement), and exact match rates should be read alongside it. Secondly, the apparent effect of question formality dissolves once infrastructure failures are excluded (Section \ref{sec:formality}). A natural next step, enabled by the failure analysis of Section \ref{sec:multidim}, is to expand the question set along the failure modes identified here rather than uniformly.

\section{Conclusion}
We conclude by stating that a careful design of LLM-as-a-Judge is necessary to obtain a reliable system that is capable of scalable and autonomous evaluation of agentic systems for drug discovery. By introducing dimensions of the design overlooked by previously proposed evaluation systems, such as human alignment validation and domain-specific judge optimization, we contribute to a more reliable evaluation of AI systems that show great promise in drug discovery. 

\subsection*{Code and Data Availability}
The ChatInvent agent evaluated in this work is available on GitHub at \\ \url{https://github.com/MolecularAI/langdmta-lab}.
For an overview of the experiments performed in this study, see \emph{evaluation/langdmta\_eval/README.md}. The judge was implemented with DSPy 3.1.3 under Python 3.12; the optimized signature retains eight labeled demonstrations. All the test questions with their evaluation context are available in the \emph{tests/} folder of the repository. 

The LLM-as-a-Judge assessments, judge signatures, and anonymized expert annotations are provided on Zenodo at \url{https://zenodo.org/records/22640840}.

\begin{credits}
\subsubsection{\ackname} EG and RM acknowledge funding provided by the Wallenberg AI, Autonomous Systems, and Software Program (WASP), supported by the Knut and Alice Wallenberg Foundation. We thank the following scientists who contributed to the annotations study and provided valuable insights and feedback: Lakshidaa Saigiridharan, Thibaud Southiratn, and Nils Dunlop.
%A bold run-in heading in small font size at the end of the paper is used for general acknowledgments, for example: This study was funded by X (grant number Y).

\subsubsection{\discintname}
EG and SG are AstraZeneca employees. The authors declare no competing interests.

\end{credits}
\bibliographystyle{splncs04}
\bibliography{references}

@article{he2026democratising,
  title={Democratising real-world drug discovery through agentic {AI}},
  author={He, Jiazhen and Lai, Helen and Saigiridharan, Lakshidaa and Ghiandoni, Gian Marco and Jenei, Kinga and Gokalp, Umur and Nukovic, Ajsa and Engkvist, Ola and Janet, Jon Paul and Genheden, Samuel},
  journal={Drug Discovery Today},
  pages={104605},
  year={2026},
  publisher={Elsevier},
  doi={10.1016/j.drudis.2026.104605}
}

@inproceedings{rios-garcia2025llmjudge,
  title = {{LLM}-as-{J}udge meets {LLM}-as-{O}ptimizer: Enhancing organic data extraction evaluations through dual {LLM} approaches},
  author = {R{\'i}os-Garc{\'i}a, Marti{\~n}o and Jablonka, Kevin Maik},
  booktitle = {AI4Mat-ICLR-2025: AI for Accelerated Materials Design Workshop, ICLR 2025},
  year = {2025},
  url = {https://openreview.net/forum?id=MjQml5U1Xq}
}

@article{bauer2025precedent,
  title={{Precedent Finder}: Locating Pareto-Optimal Reactions},
  author={Bauer, Christoph A and Kogej, Thierry and Genheden, Samuel and Norrby, Per-Ola},
  journal={Journal of Chemical Information and Modeling},
  volume={65},
  number={18},
  pages={9378--9382},
  year={2025},
  publisher={ACS Publications},
  doi={10.1021/acs.jcim.5c01797}
}

@article{loeffler2024reinvent,
  title={Reinvent 4: Modern {AI}--driven generative molecule design},
  author={Loeffler, Hannes H and He, Jiazhen and Tibo, Alessandro and Janet, Jon Paul and Voronov, Alexey and Mervin, Lewis H and Engkvist, Ola},
  journal={Journal of Cheminformatics},
  volume={16},
  number={1},
  pages={20},
  year={2024},
  publisher={Springer},
  doi={10.1186/s13321-024-00812-5}
}

@article{saigiridharan2024aizynthfinder,
  title={{AiZynthFinder 4.0}: developments based on learnings from 3 years of industrial application},
  author={Saigiridharan, Lakshidaa and Hassen, Alan Kai and Lai, Helen and Torren-Peraire, Paula and Engkvist, Ola and Genheden, Samuel},
  journal={Journal of Cheminformatics},
  volume={16},
  number={1},
  pages={57},
  year={2024},
  publisher={Springer},
  doi={10.1186/s13321-024-00860-x}
}

@article{zheng2023judging,
  title={Judging {LLM-as-a-Judge} with {MT-Bench} and {Chatbot Arena}},
  author={Zheng, Lianmin and Chiang, Wei-Lin and Sheng, Ying and Zhuang, Siyuan and Wu, Zhanghao and Zhuang, Yonghao and Lin, Zi and Li, Zhuohan and Li, Dacheng and Xing, Eric and others},
  journal={Advances in Neural Information Processing Systems},
  volume={36},
  pages={46595--46623},
  year={2023}
}

@inproceedings{jung2025trust,
  title={Trust or escalate: {LLM} judges with provable guarantees for human agreement},
  author={Jung, Jaehun and Brahman, Faeze and Choi, Yejin},
  booktitle={International Conference on Learning Representations},
  volume={2025},
  pages={3101--3125},
  year={2025}
}

@article{gottweis2025coscientist,
  title={Towards an {AI} co-scientist},
  author={Gottweis, Juraj and Weng, Wei-Hung and Daryin, Alexander and Tu, Tao and Palepu, Anil and Sirkovic, Petar and Myaskovsky, Artiom and Weissenberger, Felix and Rong, Keran and Tanno, Ryutaro and others},
  journal={arXiv preprint arXiv:2502.18864},
  year={2025},
  doi={10.48550/arXiv.2502.18864}
}

@article{m2024chemcrow,
  title={Augmenting large language models with chemistry tools},
  author={M. Bran, Andres and Cox, Sam and Schilter, Oliver and Baldassari, Carlo and White, Andrew D and Schwaller, Philippe},
  journal={Nature Machine Intelligence},
  volume={6},
  number={5},
  pages={525--535},
  year={2024},
  publisher={Nature Publishing Group UK London},
  doi={10.1038/s42256-024-00832-8}
}

@article{ding2025scitoolagent,
  title={{SciToolAgent}: a knowledge-graph-driven scientific agent for multitool integration},
  author={Ding, Keyan and Yu, Jing and Huang, Junjie and Yang, Yuchen and Zhang, Qiang and Chen, Huajun},
  journal={Nature Computational Science},
  volume={5},
  number={10},
  pages={962--972},
  year={2025},
  publisher={Nature Publishing Group US New York},
  doi={10.1038/s43588-025-00849-y}
}

@article{zhang2026molclaw,
  title={{MolClaw}: An Autonomous Agent with Hierarchical Skills for Drug Molecule Evaluation, Screening, and Optimization},
  author={Zhang, Lisheng and Wang, Lilong and Sun, Xiangyu and Tang, Wei and Su, Haoyang and Qian, Yuehui and Yang, Qikui and Li, Qingsong and Tang, Zhenyu and Sun, Haoran and others},
  journal={arXiv preprint arXiv:2604.21937},
  year={2026},
  doi={10.48550/arXiv.2604.21937}
}

@inproceedings{khattab2024dspy,
  title={{DSPy}: Compiling Declarative Language Model Calls into Self-Improving Pipelines},
  author={Khattab, Omar and Singhvi, Arnav and Maheshwari, Paridhi and Zhang, Zhiyuan and Santhanam, Keshav and Vardhamanan, Sri and Haq, Saiful and Sharma, Ashutosh and Joshi, Thomas T. and Moazam, Hanna and Miller, Heather and Zaharia, Matei and Potts, Christopher},
  journal={The Twelfth International Conference on Learning Representations},
  year={2024}
}

@article{langdmta,
  title={A Demonstration of an LLM-based Multi-agent System for Drug Discovery},
  author={Lakshidaa Saigiridharan and Helen Lai and K Jenei and Jiazhen He and Samuel Genheden},
  journal={Proc. of the 25th International Conference on Autonomous Agents and Multiagent Systems},
  year={2026},
  url={https://api.semanticscholar.org/CorpusID:288665810}
}

@article{mcnaughton2024cactus,
  title={{CACTUS}: Chemistry agent connecting tool usage to science},
  author={McNaughton, Andrew D and Sankar Ramalaxmi, Gautham Krishna and Kruel, Agustin and Knutson, Carter R and Varikoti, Rohith A and Kumar, Neeraj},
  journal={ACS Omega},
  volume={9},
  number={46},
  pages={46563--46573},
  year={2024},
  publisher={ACS Publications},
  doi={10.1021/acsomega.4c08408}
}

@inproceedings{papineni2002bleu,
  title={{BLEU}: a method for automatic evaluation of machine translation},
  author={Papineni, Kishore and Roukos, Salim and Ward, Todd and Zhu, Wei-Jing},
  booktitle={Proceedings of the 40th Annual Meeting of the Association for Computational Linguistics},
  pages={311--318},
  year={2002}
}

@inproceedings{lin2004rouge,
  title={{ROUGE}: A package for automatic evaluation of summaries},
  author={Lin, Chin-Yew},
  booktitle={Text Summarization Branches Out},
  pages={74--81},
  year={2004}
}

@article{cohen1968weighted,
  title={Weighted kappa: Nominal scale agreement with provision for scaled disagreement or partial credit.},
  author={Cohen, Jacob},
  journal={Psychological Bulletin},
  volume={70},
  number={4},
  pages={213},
  year={1968},
  publisher={American Psychological Association},
  doi={10.1037/h0026256}
}

@article{fleiss1971kappa,
  title={Measuring nominal scale agreement among many raters.},
  author={Fleiss, Joseph L},
  journal={Psychological Bulletin},
  volume={76},
  number={5},
  pages={378},
  year={1971},
  publisher={American Psychological Association},
  doi={10.1037/h0031619}
}

@article{naveed2025comprehensive,
  title={A comprehensive overview of large language models},
  author={Naveed, Humza and Khan, Asad Ullah and Qiu, Shi and Saqib, Muhammad and Anwar, Saeed and Usman, Muhammad and Akhtar, Naveed and Barnes, Nick and Mian, Ajmal},
  journal={ACM Transactions on Intelligent Systems and Technology},
  volume={16},
  number={5},
  pages={1--72},
  year={2025},
  publisher={ACM New York, NY},
  doi={10.1145/3744746}
}

@misc{gemini3modelcard,
  author       = {Google DeepMind},
  title        = {{Gemini 3.1 Pro Model Card}},
  year         = {2026},
  howpublished = {\url{https://deepmind.google/models/model-cards/gemini-3-1-pro/}},
  note         = {Accessed: 2026-05-21}
}

@article{singh2025openai,
  title={{OpenAI GPT-5} system card},
  author={Singh, Aaditya and Fry, Adam and Perelman, Adam and Tart, Adam and Ganesh, Adi and El-Kishky, Ahmed and McLaughlin, Aidan and Low, Aiden and Ostrow, AJ and Ananthram, Akhila and others},
  journal={arXiv preprint arXiv:2601.03267},
  year={2025},
  doi={10.48550/arXiv.2601.03267}
}

@misc{anthropic_claude_opus_4_7,
  author       = {Anthropic},
  title        = {{Claude Opus 4.7 Model Card}},
  year         = {2026},
  howpublished = {\url{https://www.anthropic.com/news/claude-opus-4-7}},
  note         = {Accessed: 2026-05-18}
}

@misc{llama31,
  author = {{Meta AI}},
  title = {{Llama 3.1} Model Card},
  year         = {2024},
  howpublished = {\url{https://github.com/meta-llama/llama-models/blob/main/models/llama3\_1/MODEL\_CARD.md}},
  note         = {Accessed: 2026-06-03}
}

\appendix
\setcounter{figure}{0}
\renewcommand{\thefigure}{A\arabic{figure}}

\setcounter{table}{0}
\renewcommand{\thetable}{A\arabic{table}}

\section{Question Formality Levels}

Table \ref{tab:question_variations} illustrates how the same underlying task is rephrased across the five formality levels used in our study. Four representative categories are shown: Synonyms2Smiles, DesignOnly, SynthesisOnly, and DesignSynthesis. Within each category, variation 0 is the manually-written reference question, and variations 1--4 are LLM-generated rephrasings spanning highly formal (1) through informal/slang (4) language. The placeholders \texttt{<MOLECULE\_NAME>} and \texttt{<SMILES>} are filled in with sampled values at evaluation time.

\renewcommand{\arraystretch}{1.4} % try 1.2–1.5
\begin{table}[htb!]
    \centering
    \caption{Examples of questions from four different categories and the generated variations with respect to formality. Variation 0 is the original manually-written reference question; variations 1--4 are LLM-generated rephrasings ranging from highly formal (1) to informal (4).}
    \begin{tabular}{|l|>{\raggedright\arraybackslash}p{9.5cm}|c|}
         \hline
         \textbf{Question Category} & \textbf{Question} & \textbf{Variation}  \\
         \hline
         \hline
         \multirow{5}{*}{{Synonyms2Smiles}} & "What is the SMARTS of <MOLECULE\_NAME>?" & 0 \\
         & "Could you please provide the SMARTS representation for <MOLECULE\_NAME>?" & 1 \\
         & "I need to find the SMARTS string for <MOLECULE\_NAME>, can you assist with that?" & 2 \\
         & "Hey, can you pull the SMARTS pattern for <MOLECULE\_NAME> for me?" & 3 \\
         & "whats the smarts for <MOLECULE\_NAME>??"  & 4 \\
         \hline
         \multirow{5}{*}{{DesignOnly}} & "Can you generate similar molecules to: <MOLECULE\_NAME>?" & 0 \\
         & "Would you please generate a set of molecular analogs based on the structure of <MOLECULE\_NAME>?" & 1 \\
         & "Could you generate some molecules that are structurally similar to <MOLECULE\_NAME>?" & 2 \\
         & "Hey, can you generate a few analogs of <MOLECULE\_NAME> for me?" & 3 \\
         & "can u gen some similar mols to <MOLECULE\_NAME> pls?" & 4 \\
         \hline
         \multirow{5}{*}{{SynthesisOnly}} & "How can I synthesize: <MOLECULE\_NAME> and what are some similar reactions to each reaction step involved in making this molecule?" & 0 \\
         & "Could you please propose a synthetic pathway for <MOLECULE\_NAME>? For each transformation within the proposed route, I would also require examples of analogous reactions." & 1 \\
         & "I need a viable synthetic route to <MOLECULE\_NAME>. In addition, please provide some similar reactions for each step of the synthesis." & 2 \\
         & "Can you figure out a synthesis for <MOLECULE\_NAME> for me? And for each reaction step, can you also find some other reactions that do the same kind of thing?" & 3 \\
         & "hey, how do i make <MOLECULE\_NAME>? for each step in the synth, gimme some simlar rxns too." & 4 \\
         \hline
         \multirow{5}{*}{{DesignSynthesis}} & "Can you give me some molecules similar to <SMILES> and show me the synthesis routes for the top 5 similar ones?" &  0\\
         & "I require a list of chemical analogs for the structure represented by <SMILES>. For the five most structurally similar compounds, please also provide their respective synthetic pathways." & 1 \\
         & "Could you please generate some molecules similar to <SMILES>? I would also like to see the synthesis routes for the top 5 candidates." & 2 \\
         & "Hey, I'm looking for some analogs of <SMILES>. Can you pull up a few and then show me how to make the top 5?" & 3 \\
         & "run a similarity search on <SMILES> \& show me the syntheis for the top 5 hits. thx." & 4 \\
         \hline
    \end{tabular}
    \label{tab:question_variations}
\end{table}

\section{Evaluation Dimension Correlations}
\begin{figure}[htb!]
    \centering
    \includegraphics[width=0.9\linewidth]{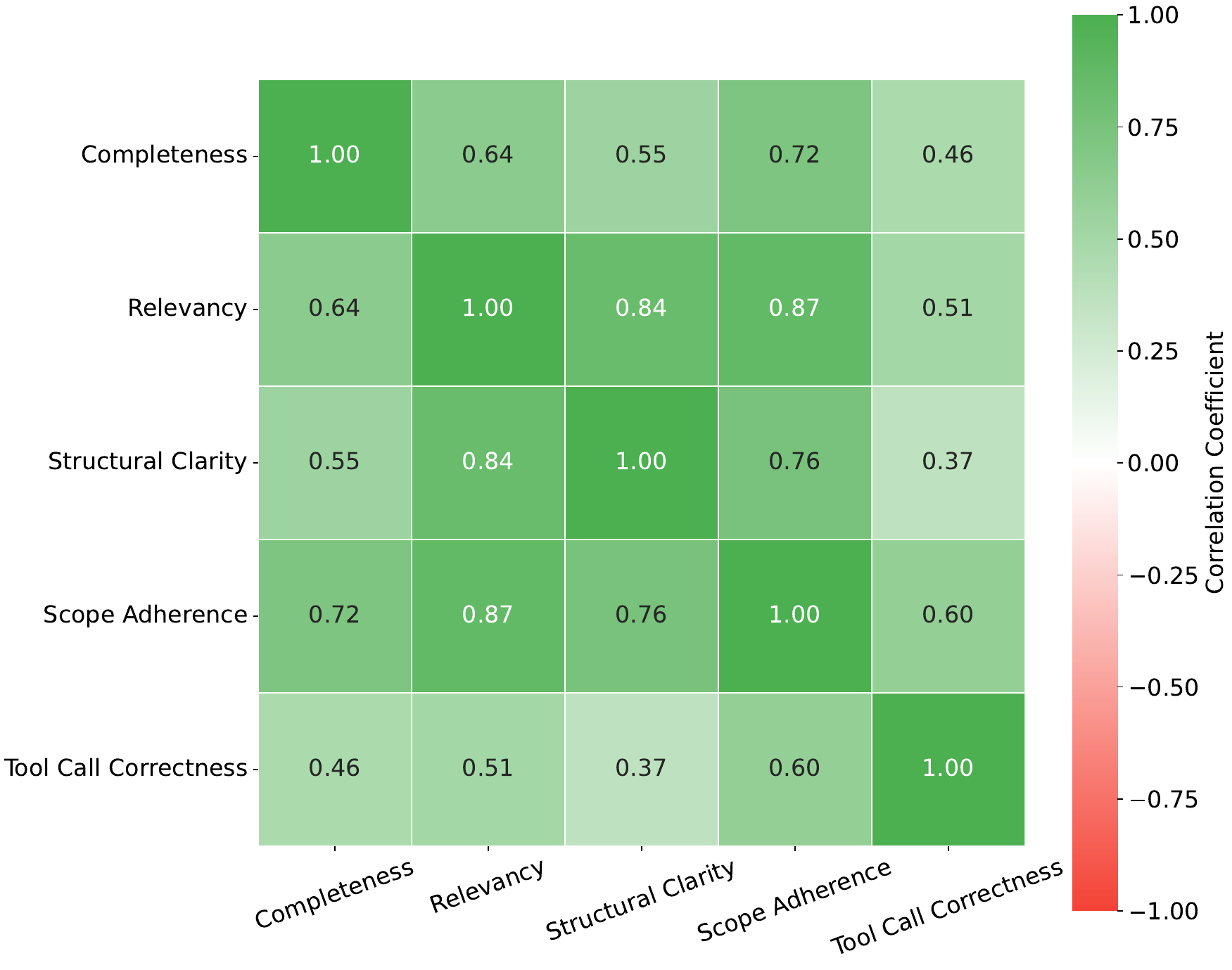}
    \caption{Correlation between evaluation dimensions across all questions in terms of Pearson correlation coefficient.}
    \label{fig:metric_correlation}
\end{figure}

In Figure \ref{fig:metric_correlation} the correlation between the evaluation dimensions. We can see that Scope Adherence and Relevancy are highly correlated and the other dimensions are all moderately correlated, based on the Pearson correlation coefficient. The high correlation between Relevancy and Scope Adherence is likely due to the nine questions that resulted in an error, as all of these scored low in both these dimensions and high on almost all other questions.

\end{document}